%% file: main_arx.tex
\documentclass{article}

\usepackage{PRIMEarxiv}

\usepackage[utf8]{inputenc} 
\usepackage[T1]{fontenc}    
\usepackage{hyperref}       
\usepackage{url}            
\usepackage{booktabs}       
\usepackage{amsfonts}       
\usepackage{nicefrac}       
\usepackage{microtype}      
\usepackage{lipsum}
\usepackage{fancyhdr}       
\usepackage{graphicx}       
\graphicspath{{media/}}     
\usepackage{amsmath} 
\usepackage{setspace}
\usepackage{graphicx} 
\usepackage{pifont}
\usepackage{makecell}
\usepackage{enumitem}
\setlist[itemize]{nosep, left=1em}
\usepackage{multirow}
\usepackage{graphicx}
\usepackage{float}
\usepackage[table]{xcolor}
\usepackage[normalem]{ulem}
\useunder{\uline}{\ul}{}

\title{Multi-Modal Monocular Endoscopic Depth and Pose Estimation with Edge-Guided Self-Supervision} 

\author{
Xinwei Ju, Rema Daher, Danail Stoyanov, Sophia Bano, Francisco Vasconcelos \\
UCL Hawkes Institute, Department of Computer Science \\
University College London, London, UK \\
\texttt{\{xinwei.ju.22, rema.daher.20, sophia.bano, danail.stoyanov, f.vasconcelos\}@ucl.ac.uk}
}

\begin{document}
\maketitle
\begin{abstract}
Monocular depth and pose estimation play an important role in the development of colonoscopy-assisted navigation, as they enable improved screening by reducing blind spots, minimizing the risk of missed or recurrent lesions, and lowering the likelihood of incomplete examinations. However, this task remains challenging due to the presence of texture-less surfaces, complex illumination patterns, deformation, and a lack of in-vivo datasets with reliable ground truth. In this paper, we propose \textbf{PRISM} (Pose-Refinement with Intrinsic Shading and edge Maps), a self-supervised learning framework that leverages anatomical and illumination priors to guide geometric learning. Our approach uniquely incorporates edge detection and luminance decoupling for structural guidance. Specifically, edge maps are derived using a learning-based edge detector (e.g., DexiNed or HED) trained to capture thin and high-frequency boundaries, while luminance decoupling is obtained through an intrinsic decomposition module that separates shading and reflectance, enabling the model to exploit shading cues for depth estimation. Experimental results on multiple real and synthetic datasets demonstrate state-of-the-art performance. We further conduct a thorough ablation study on training data selection to establish best practices for pose and depth estimation in colonoscopy. This analysis yields two practical insights: (1) self-supervised training on real-world data outperforms supervised training on realistic phantom data, underscoring the superiority of domain realism over ground truth availability; and (2) video frame rate is an extremely important factor for model performance, where dataset-specific video frame sampling is necessary for generating high quality training data.
\end{abstract}

\keywords{Monocular depth estimation \and Pose estimation \and Self-supervised learning \and Colonoscopy navigation}

\section{Introduction}


Gastrointestinal endoscopy is an essential procedure for detecting and treating cancer and other lesions in the digestive tract. However, blind spots, limited visibility, and operator variability can result in missed polyps, incomplete examinations, or higher recurrence risk. Computer-assisted navigation can address these challenges by improving lesion detection and localisation, thereby supporting more reliable and effective screening. Vision-based motion and depth estimation are key elements towards computer-assisted navigation systems that do not require additional tracking devices or any hardware modifications to a standard endoscope. However, these are still challenging tasks as endoscopy presents complex motion patterns and difficult illumination conditions. Furthermore, the lack of reliable ground truth pose and depth for real endoscopy footage makes it more difficult to train and evaluate models in this domain. As an alternative, labeled phantom or virtual endoscopy datasets are often used; however, these exhibit substantial domain shifts compared to real in-vivo endoscopy.

Both self-supervised (on real endoscopy)\cite{liu2018self,yang2024self,yang2024selflight,he2024monolot} and supervised pose and depth estimation models (on synthetic endoscopy)~\cite{jeong2021depth,jeong2024depth} have been proposed, however, both face challenges. While supervised methods are difficult to generalize to real data, self-supervised methods struggle with learning motion patterns from real scenes with abundant occlusions, illumination changes, and smooth tissue texture. Moreover, prior literature mainly focusses on model architecture innovations, while the influence of training configurations, such as supervision type (self-supervised vs. supervised), data source (real vs. phantom), and temporal sampling, have been less studied and experimentally validated.

\par
To address these challenges, we propose \textbf{PRISM} (Pose-Refinement with Intrinsic Shading and edge Maps), a stage-wise self-supervised framework that integrates additional shape and illumination cues: (1) colon fold edge maps, detected with luminace-extracted model trained on the SegCol dataset~\cite{ju2024segcol}; and (2) luminance maps~\cite{li2024image}, which help the model disambiguate shading effects from true geometry. This design helps overcome endoscopic challenges by providing clearer structural boundaries through edge maps and reducing confusion between lighting and surface shape through luminance cues, leading to more stable pose estimation and more accurate depth reconstruction. We also perform a range of comparative experiments to establish the best baseline methodology in terms of training data selection and supervision methodology. 

Our main contributions are as follows:


\begin{itemize}
\item We introduce a multi-modal self-supervised framework that integrates luminance and edge cues into depth and pose estimation, explicitly leveraging endoscopy-specific structural and illumination priors.
\item We design a stage-wise training strategy where edge maps are not only provided as input but also serve as supervisory signals through an edge similarity loss. This yields the best balance between pose and depth accuracy across datasets.
\item We provide a systematic analysis of training domain, temporal frame sampling, and weak supervision, offering new insights into how these factors influence depth and pose learning in endoscopy.
\end{itemize}

Our model outperforms the state-of-the-art on depth estimation and achieves comparable performance on pose estimation on phatom data and demonstrates improved robustness to illumination reflections and produces sharper depth contrast around fold edges on real data. Our systematic analysis further shows that: (a) training on real-world data yields better generalization than phantom or synthetic data, even on synthetic test sets; (b) optimal temporal sampling varies significantly across datasets and models; and (c) weak supervision (e.g., edge-guided loss) improves pose estimation without compromising depth accuracy.

\section{Related Work}
\label{sec:rw}
Self-supervised learning has emerged as a powerful approach for monocular depth and pose estimation without the need for ground truth labels. Methods such as SfMLearner~\cite{zhou2017unsupervised}, Monodepth and Monodepth2~\cite{godard2017unsupervised,godard2019digging}, and MonoViT~\cite{zhao2022monovit} jointly train depth and pose estimation networks by optimising photometric consistency between  different views of the same scene. Recent work has adapted these methods to endoscopy, both for monocular \cite{yang2024selflight,liu2023self} and stereo~\cite{yang2021dense} imaging modalities. Some methods explicitly incorporate endoscopy-specific visual cues into their models. Increasing robustness to geometric ambiguities caused by visual occlusions and smooth textures been considered in~\cite{huang2025occlusion} through data augmentation, in~\cite{zhu2025enhancing} through pre-trained feature representations, and in~\cite{9904879,YANG2023105989} through surface normal and optical flow priors. Strategies to minimise variance to brightness changes due to light source motion have been considered in \cite{shao2022self,zhou2023tackling}. Despite these advances, existing approaches largely overlook explicit structural priors and depend mainly on RGB inputs, leaving them vulnerable to illumination changes, specular highlights, and texture-poor regions.

Recent methods have further explored the role of illumination cues in enhancing depth prediction. Instead of treating lighting as a nuisance, these approaches leverage it to infer underlying geometry. PC-Depth~\cite{li2025unsupervised} enforces photometric consistency across frames with varying lighting conditions to recover 3D structure by modeling how illumination changes affect appearance. LightDepth~\cite{rodriguez2023lightdepth} introduces a light fall-off model and a specular reflection loss to better capture non-Lambertian surface properties, integrating light direction and intensity into the learning process while still relying on a standard depth-pose network. IID-SfMLearner~\cite{li2024image} disentangles albedo, shading, and depth via an intrinsic decomposition network, using a shading adjustment module to compensate for lighting inconsistencies during training. SHADeS~\cite{daher2025shades} extends this line by explicitly modelling non-Lambertian components through image decomposition, directly separating specular highlights from shading and reflectance without relying on correction networks. While these methods aim to filter out or correct lighting effects through decomposition or adjustment, our approach takes a complementary direction: instead of suppressing illumination variations, we embrace them by incorporating luminance decomposition as an auxiliary input, leveraging lighting as a geometric cue to enhance depth estimation. Unlike LightDepth or IID-SfMLearner, we do not introduce explicit reflectance or shading modules, nor do we seek to model specularity directly as in SHADeS. Rather, we retain luminance information to implicitly inform the structural learning process.

Beyond illumination, geometric cues such as edges also play an important role in inferring 3D structure. Learning-based Edge detection has gained popularity in computer vision with methods such as the Dense Extreme Inception Network (DEIN)~\cite{poma2020dense} and DexiNed~\cite{soria2023dense} demonstrating the ability to produce fine and perceptually plausible edge maps in natural scenes. These models have recently been adapted to colonoscopy to detect mucosal folds, using  self-supervised~\cite{jin2023self} and supervised~\cite{ju2024segcol} techniques. Scene edges have been used as additional cues for polyp segmentation~\cite{bui2024meganet}, however, their role in supporting pose and depth estimation has not yet been investigated.

While self-supervised pose and depth learning has shown promising signs in endoscopy, the limited access to real data and the lack of ground truth has motivated research towards synthetic data. Colonoscopy datasets captured with a physical phantom model~\cite{bobrow2023colonoscopy} (C3VD), virtual simulation~\cite{rau2024simcol3d} (SimCol), and in-vitro porcine organs (EndoSLAM) \cite{ozyoruk2021endoslam} have accurate pose and depth ground truth and thus have been used as training data and benchmarks for supervised and self-supervised depth and pose estimation models. Some works have focused on domain transfer techniques~\cite{jeong2024depth,mahmood2018deep,rau2023task} to make synthetic data more realistic. Notably, there has been limited investigation into comparing unlabeled real data with labeled synthetic data in terms of their effectiveness for training pose and depth models.


\section{Methodology}
\label{sec:method}
\begin{figure}[ht!]
    \centering
    \includegraphics[width=\linewidth]{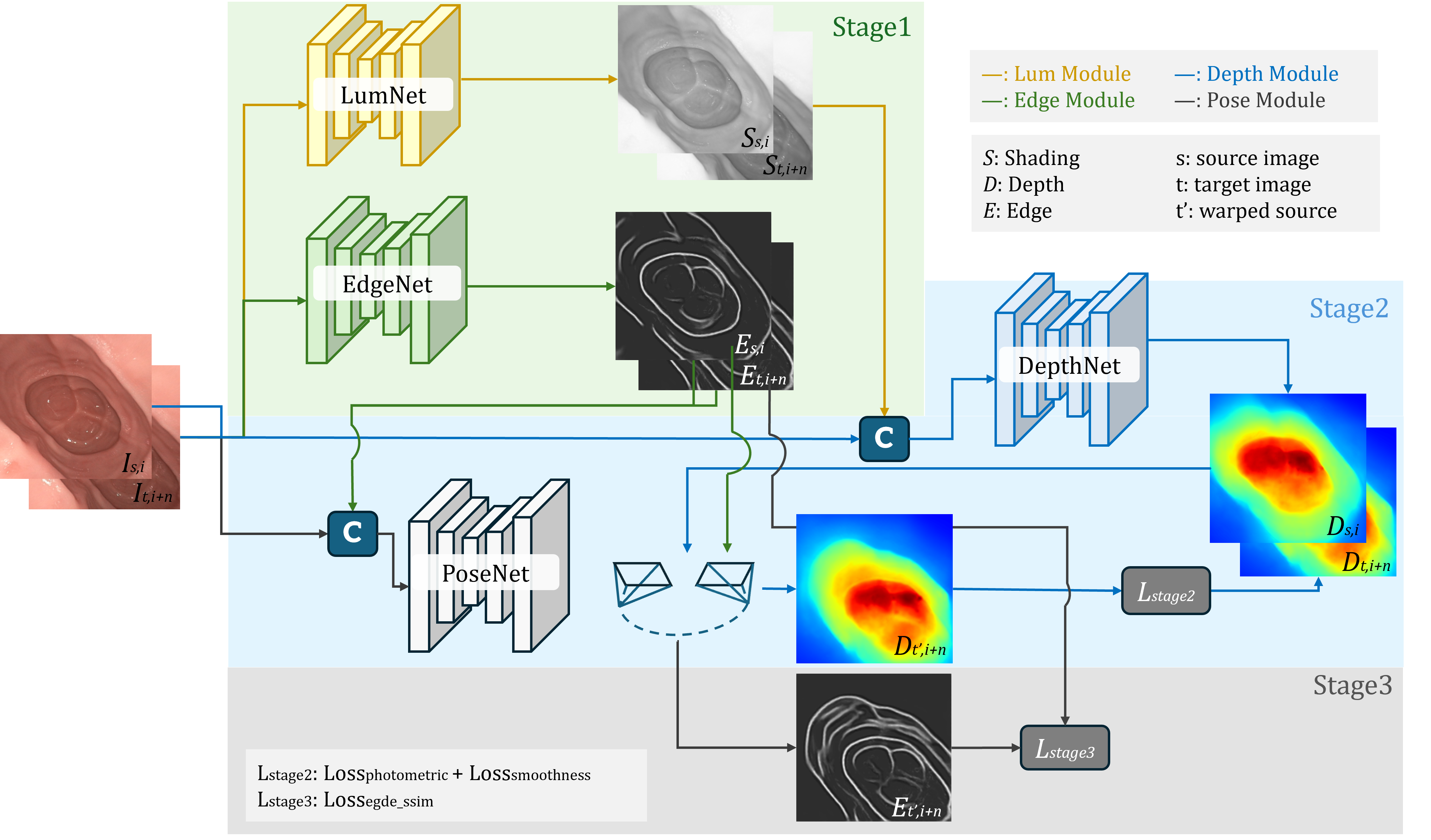}
    \caption{
        An overview of our proposed self-supervised monocular depth estimation framework. 
        Our key contribution is to explicitly guide the DepthNet with priors from two parallel branches: a LumNet, inspired by IID-SfMLearner, to extract an illumination-aware luminance map ($L_t$), and an EdgeNet, based on DexiNed, to extract a high-fidelity structural edge map ($E_t$). 
        The inputs to the DepthNet are the concatenation of the raw image $I_t$, the luminance map $L_t$, and the edge map $E_t$. 
    }
    \label{fig:framework_overview}
\end{figure}

Our method builds on self-supervised frameworks for monocular depth and pose estimation~\cite{godard2019digging}, which jointly train a DepthNet and a PoseNet from solely RGB frames. In endoscopy, this is insufficient due to illumination variation, specular highlights, and low-texture surfaces. To address these challenges, we introduce two additional networks LumNet and EdgeNet to provide structural and photometric priors. An overview of the full architecture is shown in Figure~\ref{fig:framework_overview}, and each stage and component is described in detail below.

\subsection{Network Components}

Our framework consists of four main networks: a luminance extractor (LumNet), an edge detector (EdgeNet), a depth estimator (DepthNet), and a pose estimator (PoseNet). These components provide complementary priors and predictions to support self-supervised learning under challenging endoscopic conditions.


\textbf{Luminance Extractor (LumNet).} To extract luminance maps, we utilize a model trained with the methodology from SHADeS~\cite{daher2025shades}. This is a self-supervised framework that learns different models for decoupling endoscopic images into luminance, albedo, and specularity components. In the context of this work, we only utilize luminance maps as extra features for pose and depth estimation, since in endoscopy illumination intensity is correlated with scene depth and camera motion. Importantly, this also suppresses specularities, whose dynamic and irregular patterns would otherwise reduce the robustness of depth and pose predictions.

\textbf{Edge Detector (EdgeNet).} The architecture of EdgeNet follows the same structure as the Dense Extreme Inception Network (DexiNed)~\cite{soria2023dense}, a state-of-the-art architecture for general-purpose edge detection. Dexined's publicly released pre-trained model is very sensitive to light reflections abundant in endoscopy. Therefore, we train EdgeNet from scratch in a fully supervised manner on real colonoscopy data labeled with fold segmentation maps from the MICCAI Endovis SegCol Challenge dataset~\cite{ju2024segcol}.

\textbf{Pose and Depth Networks.} Following standard practice in self-supervised methods~\cite{godard2017unsupervised}, we utilize two separate networks (PoseNet and DepthNet) to estimate camera pose and scene depth. While prior methods typically feed raw RGB frames into both networks, we introduce additional structural and photometric priors to improve performance under challenging conditions. The PoseNet is a convolutional network that predicts the relative 6-DoF camera pose between a pair of frames ($I_t$, $I_s$ $\rightarrow \hat{T}_{t \to s}$). The input to this network is a concatenation of frames $(I_t,I_s)$ and scene edge detections $(E_t, E_s)$ which aid in view alignment. The DepthNet adopts a standard encoder-decoder architecture from~\cite{godard2019digging}. The input to this network is the concatenation between frames $(I_t,I_s)$ and luminance maps $L_t, L_s$, which help accounting for the correlation between scene depth and illumination intensity.   

\subsection{Training Strategy}
\label{sec:trainingstrategy}

We propose a self-supervised framework for monocular depth estimation from video sequences. As illustrated in Figure~\ref{fig:framework_overview}, the architecture includes four encoder-decoder modules: a Luminance extractor (\textbf{LumNet}), an Edge detector (\textbf{EdgeNet}), a camera pose estimator (\textbf{PoseNet}), and a depth estimator (\textbf{DepthNet}). These modules are trained in a three stage process.

\paragraph{Stage 1: Pre-training of LumNet and EdgeNet}  
LumNet and EdgeNet produce additional illumination and anatomy cues to be used by pose and depth estimation models (PoseNet, DepthNet). These models are pre-trained and their weights are frozen in subsequent training stages.
  
\paragraph{Stage 2: Joint Training of PoseNet and DepthNet}  
DepthNet and PoseNet are jointly trained using a self-supervised photometric loss in line with state-of-the-art methods such as MonoDepth2 \cite{godard2019digging}. Differently from previous methods, our model utilizes Edge and Luminance maps in addition to images as input to depth and pose models.


\paragraph{Stage 3: PoseNet Refinement with Edge-Aware Loss.}  
We experimentally verify that stage 2 training, compared to baselines, generally improves the accuracy of DepthNet at the detriment of worsening PoseNet results. Therefore, we further finetune PoseNet (with DepthNet frozen) using an additional loss term that measures geometric consistency between edge maps of view pairs. 


\subsection{Self-Supervised Loss Functions}
\label{sec:loss_functions}

DepthNet and PoseNet are jointly trained with a self-supervised objective that enforces geometric consistency between adjacent video frames. All loss terms are computed in a multi-scale fashion across the depth decoder outputs, following~\cite{godard2019digging}. 
\begin{itemize}
    \item \textbf{Photometric Reprojection ($L_{\text{stage2}}$)}:
    In stage 2, DepthNet and PoseNet are jointly trained using a combination of photometric reprojection loss and gradient-guided smoothness loss:
    \begin{equation}
    L_{\text{stage2}} = \sum_{\sigma=0}^{3} \left[ \lambda_{\text{photo}} \min_{t'} pe(I_t^\sigma, I_{t' \rightarrow t}^\sigma) + \lambda_{\text{smooth}} \left( |\partial_x d_t^\sigma| e^{-|\partial_x I_t^\sigma|} + |\partial_y d_t^\sigma| e^{-|\partial_y I_t^\sigma|} \right) \right]
    \end{equation}
    Here, $\sigma \in \{0,1,2,3\}$ indexes the multi-scale levels; $I_t^\sigma$ and $d_t^\sigma$ denote the target image and inverse depth, while $\partial_x$, $\partial_y$ denote horizontal and vertical gradients. The warped image $I_{t' \rightarrow t}^\sigma$ is computed using predicted depth and relative pose. The photometric error is defined as:
    \begin{equation}
    pe(I_a, I_b) = \frac{\alpha}{2}(1 - \text{SSIM}(I_a, I_b)) + (1 - \alpha)\|I_a - I_b\|_1
    \end{equation}
    with $\alpha = 0.85$, following~\cite{godard2019digging}. We apply the minimum photometric error across all source views to suppress supervision from occluded or misaligned frames, improving training robustness. The smoothness term encourages spatial coherence in inverse depth while preserving edges. We set $\lambda_{\text{photo}} = 1.0$ and $\lambda_{\text{smooth}} = 0.1$.

 \item \textbf{Edge-Guided Structural Consistency Loss ($L_{\text{stage3}}$)}:  
    To improve geometric alignment across views, we fine-tune the PoseNet using an edge-guided structural consistency loss. This loss is introduced after an initial training stage where DepthNet and PoseNet are jointly optimized using the photometric and smoothness losses in $L_{\text{stage2}}$. The overall stage 3 objective is:
    \[
    L_{\text{stage3}} = L_{\text{stage2}} + \lambda_{\text{edge}} L_{\text{edge}}
    \]
    with $\lambda_{\text{edge}} = 1.0$ for equal weighting. For each source frame $t'$ and each scale $\sigma \in \{0,1,2,3\}$, we extract its edge map $E_{t'}^\sigma$ and warp it to the target view using the predicted depth $D_t^\sigma$ and relative pose $T_{t \rightarrow t'}$. The warped map $E_{t' \rightarrow t}^\sigma$ is compared with the target edge map $E_t^\sigma$ using SSIM:
    \begin{equation}
    \label{eq:edgeloss}
    L_{\text{edge}} = \sum_{\sigma=0}^{3} \frac{1}{N^\sigma} \sum_x \left( 1 - \text{SSIM}(E_t^\sigma(x), E_{t' \rightarrow t}^\sigma(x)) \right)
    \end{equation}
    where $N^\sigma$ is the number of valid pixels at scale $\sigma$.
    
    This loss shares the same warping transformation and SSIM formulation as the photometric term in $L_{\text{stage2}}$, but operates on edge maps to emphasize geometric alignment of object boundaries. While $L_{\text{stage2}}$ promotes appearance consistency across views, the edge-guided loss complements it by enforcing structural consistency, leading to depth and pose predictions that are both visually coherent and structurally precise across scales.

\end{itemize}

\section{Experiments and Results}
\label{sec:experiments}


In this section, we firstly introduce the experimental setup (Sec.~\ref{sec:setup}).
We then compare our final model against state-of-the-art methods, presenting both quantitative and qualitative results (Sec.~\ref{sec:main_results}).
To validate our specific design choices, we also present a dedicated ablation study on the contributions of our proposed multi-modal components (Sec.~\ref{sec:ablation_mod}) and the effectiveness of our stage-wise loss design under structural guidance (Sec.~\ref{sec:ablation_loss}). We refer the reader to Sec.~\ref{sec:appendix_analysis} for an extended analysis and summary of training domain, temporal sampling, and supervision strategy.

\subsection{Experimental Setup}
\label{sec:setup}

\subsubsection{Datasets}


\textbf{Hyper-Kvasir:} A dataset of real endoscopy short video clips~\cite{borgli2020hyperkvasir}. We utilize it for training all self-supervised models (LumNet,PoseNet,DepthNet). We utilized the subset of videos with good bowel preparation (BBPS 2-3), and sampled the video at 25 fps, with 16,976 frames for training. 

\textbf{C3VD:} A realistic phantom colonoscopy dataset with pose and depth ground truth~\cite{bobrow2023colonoscopy}, recorded at 30 fps (frames per second). We utilize its test split for quantitative evaluation and its training split for ablation experiments comparing supervised and self-supervised models. We specifically report performance on four held-out trajectories: \texttt{cecum\_t4b}, \texttt{descending\_t4a}, \texttt{sigmoid\_t3b}, and \texttt{transcending\_t4b}, 1706 frames in total, following the dataset split in~\cite{he2024monolot}.

\textbf{EndoMapper:} A large dataset with complete real endoscopy videos~\cite{azagra2023endomapper}. We utilize it for qualitative model evaluation, using frames from sequences 1, 16, and 95.

\textbf{SegCol:} A subset of the EndoMapper dataset, with manually labeled edge contours of colon folds, as part of the SegCol Challenge~\cite{ju2024segcol}. We utilize the same training and validation splits as the SegCol challenge to train EdgeNet. 


We note that while the C3VD dataset offers dense ground truth for synthetic colonoscopy scenes, our main model is trained on Hyper-Kvasir instead. This choice is based on empirical evidence (see Sec.~\ref{sec:appendix_analysis}) that real endoscopy training data leads to more accurate results, even when evaluated on C3VD synthetic data. 

\subsubsection{Hyper-parameters}

Our framework is implemented in PyTorch and all experiments were conducted on a single NVIDIA A100-SXM GPU. We employ the Adam optimizer with default parameters ($\beta_1 = 0.9$, $\beta_2 = 0.999$). The initial learning rate is set to $1 \times 10^{-4}$ and is reduced by a factor of 0.1 after 15 epochs using a step-based scheduler. According to individual intrinsics, input images are undistorted, cropped, and resized to $288 \times 288$ pixels before being fed into the networks. A batch size of 12 is used for all training stages.


In stage 1, we trained EdgeNet for 20 epochs. For LumNet, we utilized pre-trained weights from \cite{daher2025shades} (which were also trained for 20 epochs). In stages 2 and 3, we trained PoseNet and DepthNet for 20 epochs in each stage. In Stage 2, PoseNet and DepthNet were initialized with pre-trained weights from Monodepth2~\cite{godard2019digging}, trained on the KITTI dataset.


\subsubsection{Evaluation Metrics}

We adopt standard depth error metrics established by Eigen et al.~\cite{eigen2014depth} (Abs Rel, Sq Rel, RMSE, RMSE log, $\delta < 1.25^i$ for $i \in \{1,2,3\}$) . Additionally, we report the Mean Absolute Error (MAE) and Median Absolute Error (MedAE). For camera pose estimation, we follow the protocol from Rau et al.~\cite{rau2024simcol3d} and use the EVO toolkit~\cite{grupp2017evo} to report the Absolute Trajectory Error (ATE), Relative Trajectory Error (RTE), and Relative Orientation Error (ROT).

\input{TAB/Table_main}

\begin{figure}[ht!]
    \centering
    \includegraphics[width=\linewidth]{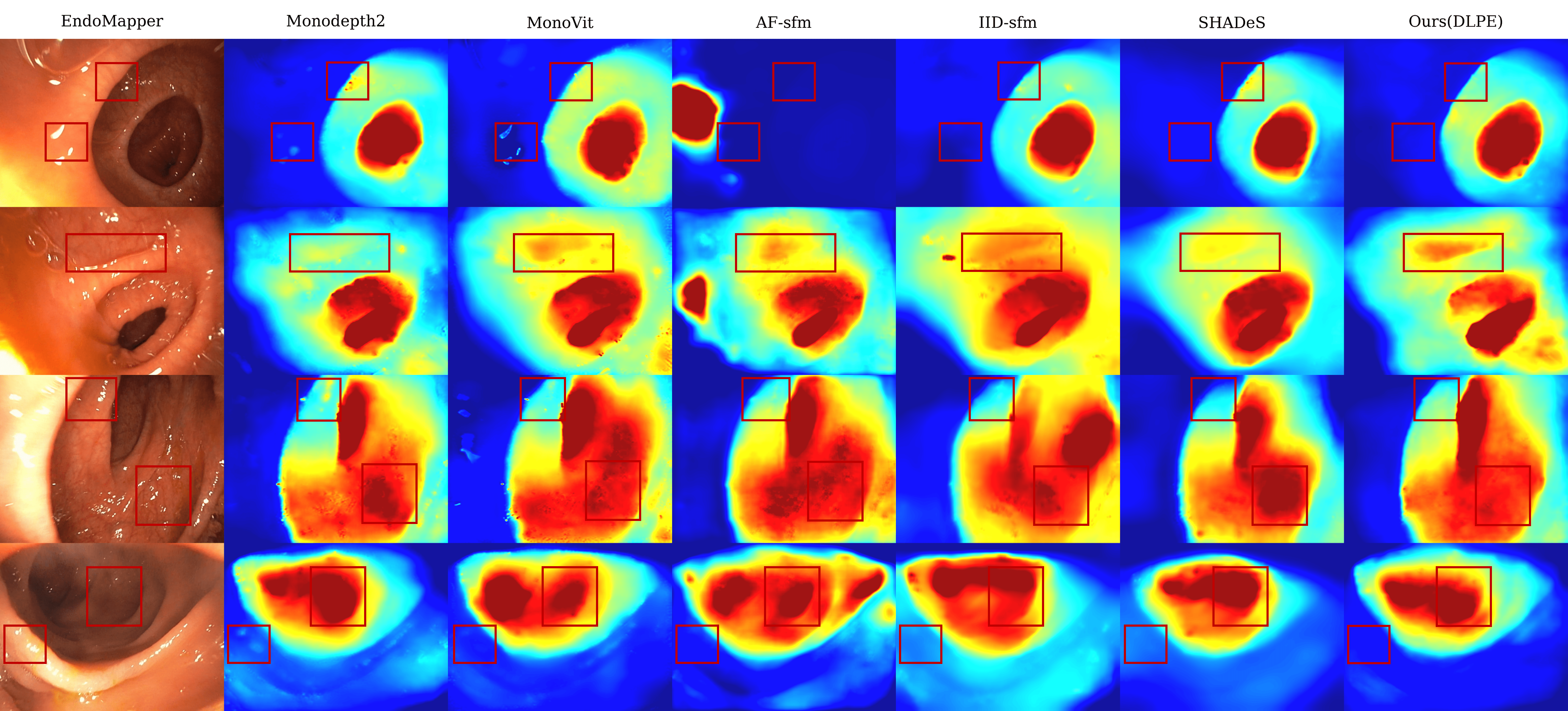}
    \caption{Qualitative evaluation of generalization on \textbf{real-world endoscopic video frames} with challenging improvements boxed in red.}
    \label{fig:endomapper_results}
\end{figure}

\begin{figure}[t]
    \centering
    \includegraphics[width=\linewidth]{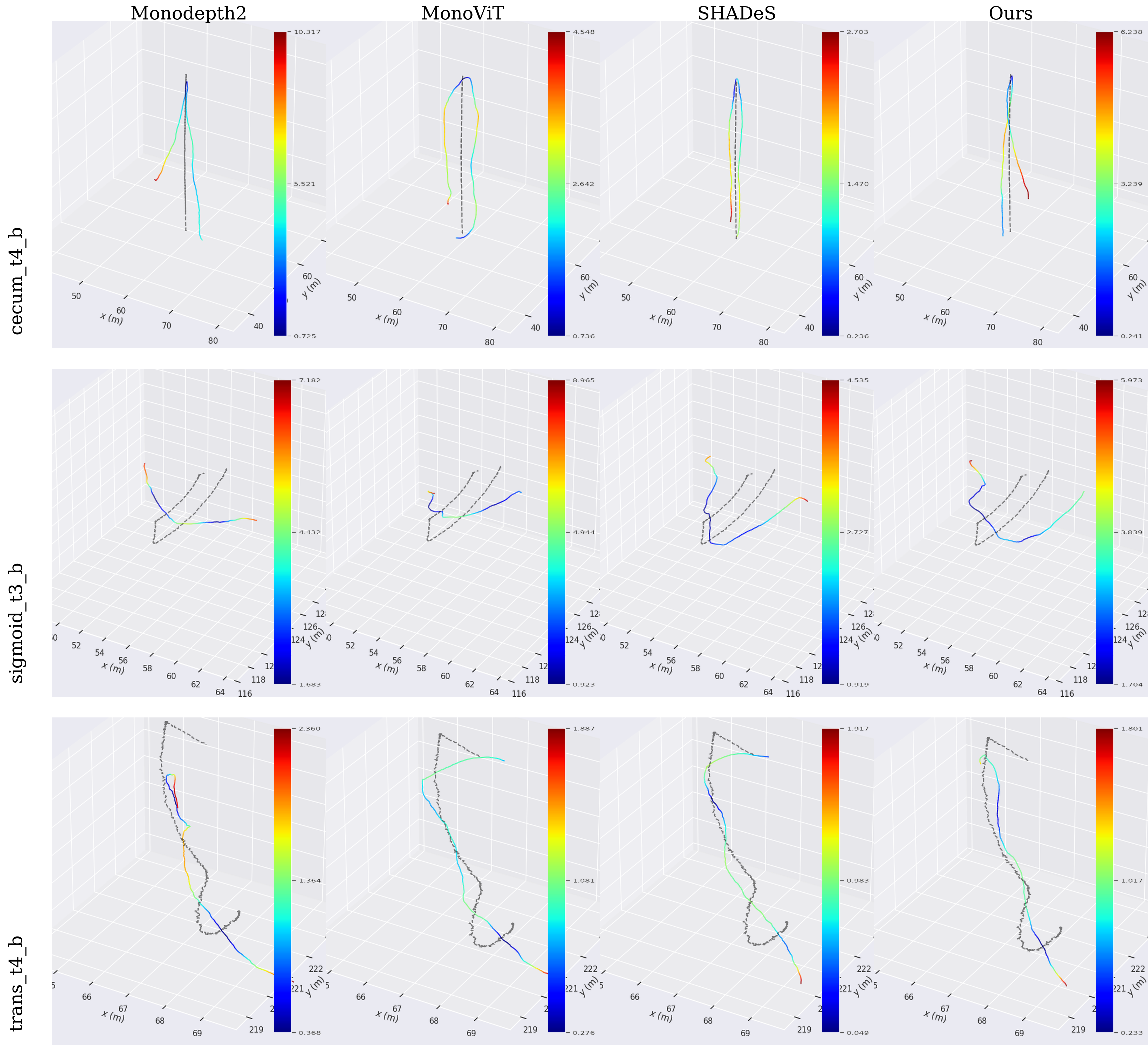}
    \caption{Qualitative comparison of estimated camera trajectories across four models on three representative sequences from the C3VD dataset. Each 3D plot shows the predicted path color-coded by temporal order.}
    \label{fig:stacked_ape_maps_zoomin_0730}
\end{figure}

\subsection{Comparison with State-of-the-Art}
\label{sec:main_results}

Our model is compared against both general-purpose~\cite{godard2019digging,zhao2022monovit} and endoscopy-specific approaches~\cite{shao2022self,li2024image,daher2025shades}. For the latter,~\cite{shao2022self, li2024image} were originally trained on a different endoscopic domain (SCARED dataset~\cite{allan2021stereo}) and we evaluate both the original weights (pre-trained) and a fine-tuned version on Hyper-Kvasir (HK). \cite{daher2025shades} is directly trained on Hyper-Kvasir (HK) so we report a single version of this model.

Table~\ref{tab:depth_comparison_main} shows that our method and MonoViT are the top performers in depth estimation on C3VD synthetic data. Our model performs best in metrics sensitive to large errors (e.g. RMSE), while MonoViT performs best in metrics without this characteristic (e.g. MedAE, MAE). However, MonoViT performs significantly worse than most other models on real data (Fig.~\ref{fig:endomapper_results}), showing high sensitivity to light reflections, often producing undesirable artifacts. On the other hand, \texttt{SHADeS} also shows good robustness to reflections. Nonetheless, our method still produces sharper edges around colon folds (second row) and does not "hallucinate" lumen features in incorrect regions (third row). While our method performs well on both synthetic (C3VD) and real (EndoMapper) data, baselines show mixed results. Given the visual complexity of real endoscopy, we argue that strong performance on real data, despite lacking ground truth, should be prioritized.

The ATE results in Table~\ref{tab:depth_comparison_main} indicate that SHADeS achieves the lowest ATE on phantom data, with our method ranking second. As shown in Figure~\ref{fig:stacked_ape_maps_zoomin_0730}, both SHADeS and our model produce trajectories that are noticeably closer to the ground truth than those of other baselines. Consistent with the depth estimation results, we observe that pose estimation performance on synthetic data can differ from that on real data. However, unlike depth estimation, qualitative evaluation of real trajectories is less feasible.


In summary, our method and SHADeS provide the most convincing results across all experiments, with our method performing better in depth estimation while SHADeS performing better at pose estimation. 

\subsection{Ablation Study}
\subsubsection{Luminance and Edge Features}
\label{sec:ablation_mod}

\input{TAB/Table_modality}

\begin{figure}
    \centering
    \includegraphics[width=\linewidth]{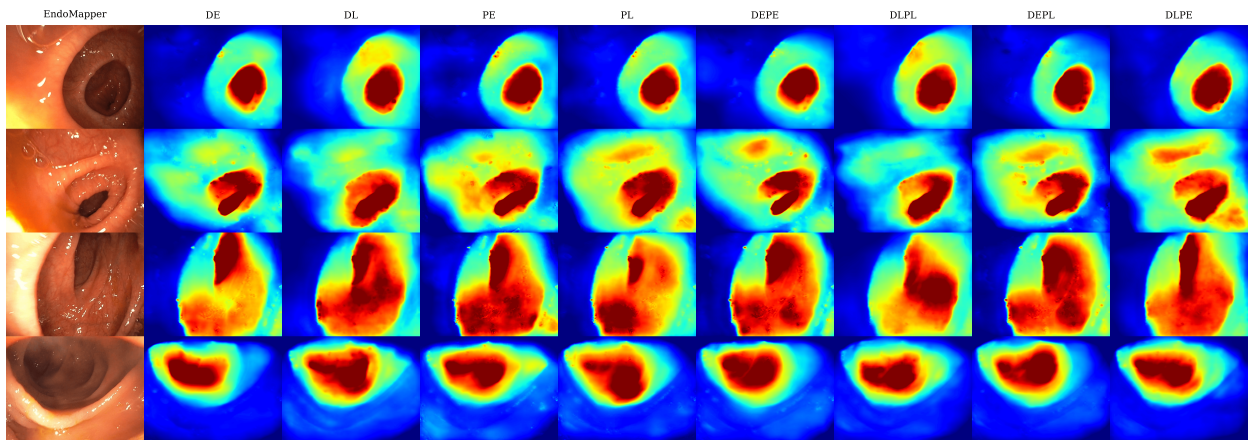}
    \caption{Qualitative comparison of ablation variants on four subsequences from the EndoMapper dataset. Each row shows a sample frame and the corresponding predicted depth maps from different prior configurations.}
    \label{fig:ablation_qualitative}
\end{figure}

We evaluate the role of luminance and anatomy edge features by providing them as input to DepthNet and PoseNet in different combinations. Table~\ref{tab:depth_ablation_hk} shows depth estimation ablation results on C3VD, where the \texttt{DLPE} configuration denotes our proposed model. Best results are obtained by providing Luminance and Edge features to DepthNet and PoseNet respectively (\texttt{DLPE}) or by providing luminance features to both models (\texttt{DLPL}). Some variants either degrade performance or show no clear benefit, indicating that indiscriminate assignment is ineffective. We also show qualitative ablation results on EndoMapper Real data in Fig.~\ref{fig:ablation_qualitative}, which shows \texttt{DLPE} to outperform \texttt{DLPL} in terms of fewer light artifacts (first row), more well defined edge contours (second row), and correct lumen location (third row). We also note some performance differences between quantitative results on phantom (C3VD) and real (EndoMapper). By prioritising results on Real-Data, we conclude \texttt{DLPE} is overall a more robust configuration than \texttt{DLPL}.



\subsubsection{Refinement with Edge-Aware Loss}
\label{sec:ablation_loss}
We assess the effect of the Edge-aware loss (Eq.~\ref{sec:trainingstrategy}) in Table~\ref{tab:mod_loss_ablation}. For the baseline (Monodepth) and the best performing configurations from Sec.~\ref{sec:ablation_mod} (DLPE, DLPL) we compare the following:
\begin{itemize}
\item No Edge-aware loss
\item Joint training of PoseNet and DepthNet with Edge-aware loss;
\item Refining PoseNet alone with Edge-aware loss in stage 3 (our proposed strategy);
\end{itemize}

\begin{figure}[ht]
\centering
\includegraphics[width=0.6\linewidth]{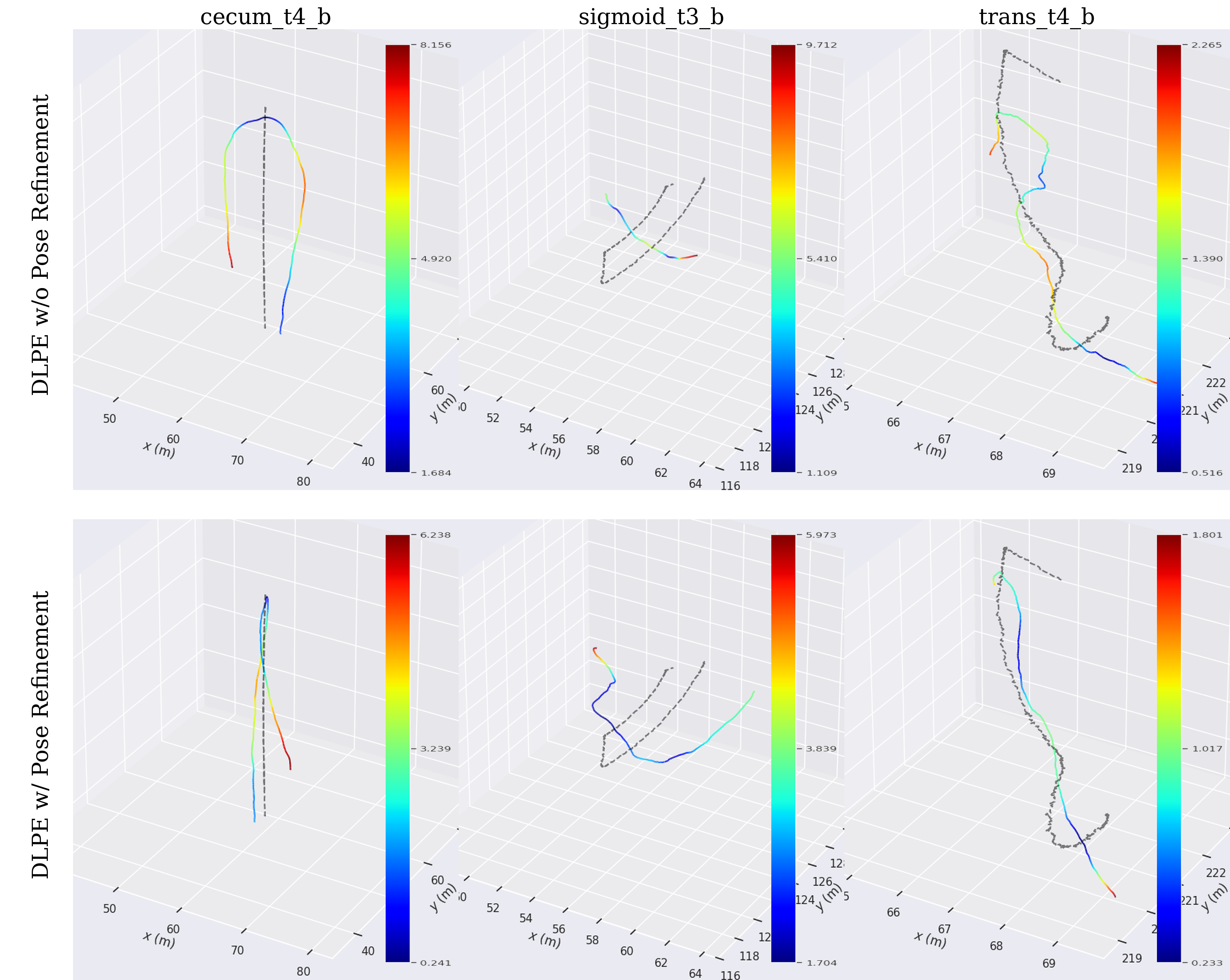}
\caption{Qualitative trajectory comparison under modality-only vs. modality + edge-guided SSIM training on \texttt{cecum\_t4b}, \texttt{sigmoid\_t3b} and \texttt{trans\_t4b}.}
\label{fig:traj_loss_vis}
\end{figure}

Stage 3 consistently improves camera trajectory reconstruction (lower ATE), however, it degrades depth estimation results if DepthNet is jointly trained in stage 3. We therefore conclude the best strategy is to freeze DepthNet in stage 3 so that we can obtain improvements in trajectory reconstruction without degrading depth estimation. Figure~\ref{fig:traj_loss_vis} further highlights visually the improvement in trajectories when refining stage 3 with Edge-aware loss.

\input{TAB/Table_mod_loss}

\subsection{Training Data Domain, Sampling, and Supervision}
\label{sec:appendix_analysis}

\begin{figure}[t]
    \centering
    \includegraphics[width=\linewidth]{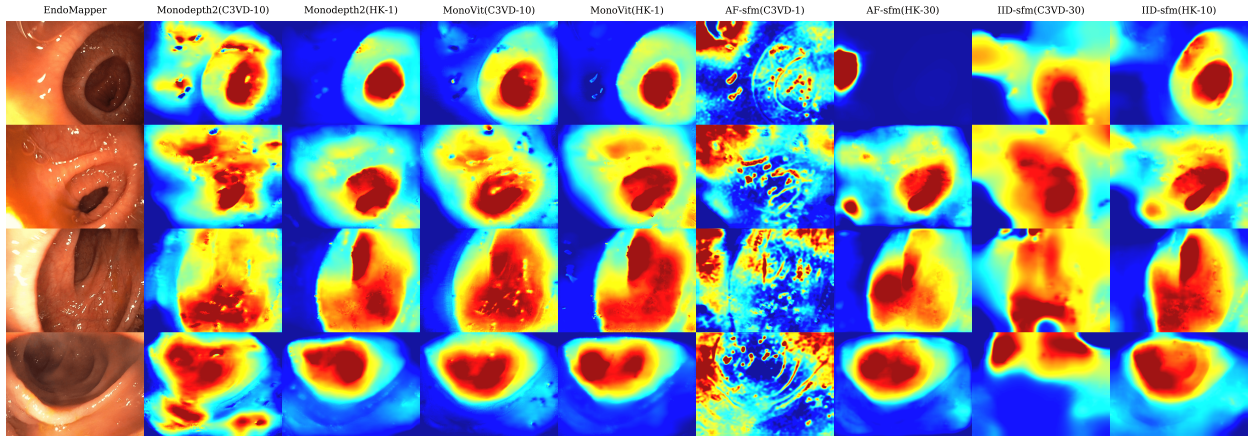}
    \caption{Qualitative generalization performance of models trained with Hyper-Kvasir on EndoMapper real dataset. This evaluation highlights model robustness to real-world challenges like specular reflections and complex tissue folds, where Hyper-Kvasir-trained models consistently show an advantage.}
    \label{fig:endomapper_dataset_temporal}
\end{figure}

\textbf{Domain and Sampling:} We further assess the influence of training data domain and sampling as critical factors for good performance that have been overlooked in prior work. Table~\ref{tab:temporal} reports depth accuracy on C3VD using models trained on either C3VD or Hyper-Kvasir. We also evaluate the effect of subsampling video frame rates during training. Surprisingly, training on Hyper-Kvasir (real data) consistently outperforms training on C3VD (phantom), even though evaluation is performed on the C3VD test data, underscoring the value of real data. Beyond visual domain differences, the datasets vary in endoscope motion; C3VD features slower, smoother movements, which may be too subtle for PoseNet to effectively learn motion cues. As shown in Table~\ref{tab:temporal}, effective training on C3VD requires subsampling by a factor of 10, while no subsampling is required for Hyper-Kvasir training. Even with optimal subsampling for both, Hyper-Kvasir training yields significantly better results. All these results are further highlighted visually in Figure~\ref{fig:endomapper_dataset_temporal}.

This suggests that inter-frame motion magnitude is a dominant factor in self-supervised depth learning. When motion is too small, the photometric constraint provides weak geometric supervision, limiting the ability of PoseNet to learn meaningful correspondences. In contrast, datasets with richer motion statistics provide stronger multi-view signals, leading to more stable convergence.

\textbf{Supervision:} Since C3VD contains reliable depth ground truth, we also explore a supervised training strategy using the optimal subsampling rate (10). Specifically, we augment the baseline with a supervised loss, using optimally tuned weights:
\begin{equation}
    L_{\text{stage2}} =  L_{\text{photo}} + 0.1 \times L_{\text{smooth}} + 0.5 \times L_{\text{sup}},
\end{equation}
\noindent
where $L_{\text{sup}}$ is the L1 distance between predicted and ground truth depth.

Surprisingly, adding supervised depth loss degrades performance across all metrics compared to the purely self-supervised baseline as shown in Table~\ref{tab:supervised_vs_self}. Although the weight on $L_{\text{sup}}$ is moderate, it can dominate in low-texture regions with weak photometric gradients, leading to memorization of noisy targets rather than learning multi-view consistency. This is partly due to the limitations of C3VD’s ground truth, obtained via structured light scanning and trajectory alignment. Although scale-consistent, the depth maps still contain noise and lack valid information in occluded regions.

To provide a consolidated comparison, Table~\ref{tab:depth_interval_summary} summarises depth estimation performance across training domains, temporal intervals, and supervision modes under identical DLPE settings. The results show that increasing the temporal interval on C3VD substantially improves performance, highlighting the importance of sufficient inter-frame motion for self-supervised learning. Training on Hyper-Kvasir achieves the best overall accuracy despite cross-dataset evaluation, underscoring the benefit of realistic motion and visual diversity. In contrast, adding supervised depth loss on C3VD does not outperform the optimally sampled self-supervised setting, suggesting that multi-view consistency remains more effective than imperfect ground-truth supervision in this scenario. In such regions, the supervised signal can dominate over the photometric term, encouraging the network to fit noisy or misaligned targets rather than enforcing geometric consistency across views.

\textbf{Summary:} Overall, Pose and Depth estimation perform best when trained self-supervised on real data. Despite being the best available endoscopy dataset with depth and pose labels, C3VD has notable limitations due to its phantom-based nature. This further highlights that scale-consistent ground truth alone is insufficient if it contains structural noise or incomplete depth observations, particularly in occluded regions.

\input{TAB/Table_temporal}


\begin{table}[t]
\centering
\caption{Comparison between self-supervised and supervised Monodepth2~\cite{godard2019digging} models on C3VD (interval = 10). Supervised model adds scale-consistent depth loss with a weight of 0.5. The better one is marked in \textbf{Bold}}
\label{tab:supervised_vs_self}
\setlength{\tabcolsep}{4pt}
\renewcommand{\arraystretch}{1.2}
\resizebox{\textwidth}{!}{
\begin{tabular}{lcccccc|ccc}
\toprule
\textbf{Model} & \textbf{RMSE} ↓ & \textbf{LogRMSE} ↓ & \textbf{MAE} ↓ & \textbf{MedAE} ↓ & \textbf{AbsRel} ↓ & \textbf{SqRel} ↓ & \textbf{$\delta<1.25$} ↑ & \textbf{$\delta<1.25^2$} ↑ & \textbf{$\delta<1.25^3$} ↑ \\
\midrule
Self-supervised & \textbf{7.5960} &\textbf{ 0.2215} & \textbf{5.0917} & \textbf{3.4709} & \textbf{0.1732} & \textbf{1.4608} & \textbf{0.7161} & \textbf{0.9253} & \textbf{0.9827} \\		
Supervised (+GT) & 8.2805 & 0.2443 & 5.6020 & 3.7030 & 0.1921 & 1.7754 & 0.6797 & 0.9022 & 0.9721
\\
\bottomrule
\end{tabular}}
\end{table}

\begin{table}[t]
\centering
\caption{Comparison of depth estimation across datasets, temporal intervals, and supervision modes using PRISM with \texttt{DLPE} settings.}
\label{tab:depth_interval_summary}
\setlength{\tabcolsep}{0.8pt}
\renewcommand{\arraystretch}{1}
\footnotesize
\begin{tabular*}{\columnwidth}{@{\extracolsep{\fill}} llccccccc}
\toprule
\textbf{Model} & \textbf{Training Set} & \textbf{Interval} & 
\textbf{RMSE} ↓ & \textbf{LogRMSE} ↓ & \textbf{MAE} ↓ & \textbf{MedAE} ↓ & 
\textbf{AbsRel} ↓ & \textbf{SqRel} ↓ \\
\midrule
\multirow{4}{*}{\textbf{\textbf{PRISM}}} 
 & C3VD & 1  & 14.4647 & 0.4165 & 9.7689 & 6.3846 & 0.3221 & 4.7946 \\
 & C3VD & 30 &  7.7530 & 0.2460 & 5.4256 & 3.9831 & 0.1972 & 1.6139 \\
 & C3VD\_supervised & 30 & 8.2836 & 0.2523 & 5.6774 & 3.8749 & 0.1976 & 1.7205 \\
 & Hyper-phasir & 1 & \textbf{6.1246} & \textbf{0.1830} & \textbf{4.3300} & \textbf{3.2582} & \textbf{0.1475} & \textbf{1.0010} \\
\bottomrule
\end{tabular*}
\end{table}

\vspace{0.5em}
\noindent

\section{Discussion of Results}
\label{sec:discussion}

Our proposed model, which integrates luminance and edge features, delivers more robust depth estimation than most baseline methods. It consistently produces depth maps with fewer artifacts and a more accurate lumen characterisation, an essential capability for endoscope guidance. While luminance and edge features bring more moderate gains in pose estimation, the recent SHADeS method, which also leverages luminance, currently leads in this task. Nevertheless, our results suggest that these features independently improve models, making a future integration of the two approaches a promising direction.

Our ablation study confirms that both edge and luminance features enhance depth estimation, albeit at the cost of reduced pose estimation accuracy. We achieve state-of-the-art level pose estimation only through an Edge-Aware refinement stage. Notably, luminance features are best exploited by the DepthNet encoder, while edge features are most beneficial in the PoseNet encoder, though both feature types jointly contribute to improved depth estimation in the joint training stage.

Across all experiments, we find significant differences between phantom (C3VD) and real (Hyper-Kvasir) datasets in both training and testing. Models performing well on C3VD often fail to generalize to real data (e.g., MonoVIT), likely because real data contains richer visual features and more complex motion patterns. We also identify video subsampling as a key factor influencing performance, with its optimal setting depending heavily on the motion characteristics of the training data.

\section{Conclusion}

We introduce a framework for self-supervised monocular depth and pose estimation in endoscopy that incorporates additional luminance and edge features. We conduct an exhaustive analysis of their roles, finding that both have a measurable impact on performance. Luminance cues, when integrated into the depth network as input features, enhance lumen characterisation and help reduce artifacts in depth maps. Edge contours, on the other hand, can be leveraged both as input features and as an auxiliary loss term, with the former being more beneficial for depth estimation and the latter for pose estimation.

Through extensive experimental comparisons, we observe marked differences in model performance between phantom and real datasets. This is a critical finding, as C3VD remains the most widely used benchmark for quantitative evaluation in colonoscopy, yet results on C3VD do not always transfer to real-world settings. We argue that both quantitative and qualitative evaluations are essential for understanding this problem, as certain challenges only emerge when testing on real data.

We also draw attention to often overlooked aspects of training methodology, such as video data sampling. Sampling strategy can significantly influence the performance of depth and pose models, and its optimal configuration depends on both the dataset and the model. This makes it a key factor to consider in the design and release of future endoscopy datasets.

Our findings motivate several avenues for future research. These include exploring alternative mechanisms for incorporating structural features into endoscopic pose and depth estimation, such as attention-based strategies or dynamic feature weighting.  Another direction is developing domain adaptation techniques to reduce dependence on real data. We also see potential in extending the approach to stereo or dynamic video input, which could further improve generalization and robustness in real-world navigation systems. Finally, soft or probabilistic supervision may offer a more stable alternative to training with noisy ground-truth annotations.



\section*{Declarations}
\textbf{Acknowledgments:} This work was supported by the Optical and Acoustic Imaging for Surgical and Interventional Sciences hub [UKRI145] and the Royal Academy of Engineering under the Chair in Emerging Technologies programme. \textbf{Conflict of interest:} Authors declare that they have no conflict of interest. \textbf{Ethical approval:} This article does not contain any studies with human participants or animals. 

\bibliography{main_arx}
\bibliographystyle{unsrt}

\end{document}

%% file: TAB/Table_main.tex
\begin{table}[t]
\centering
\caption{Quantitative comparison of depth estimation on the C3VD test set for models trained with Hyper-Kvasir. Best results are in \textbf{bold}, second best are \underline{underlined}.}
\setlength{\tabcolsep}{4pt}
\label{tab:depth_comparison_main}
\renewcommand{\arraystretch}{1.2}
\resizebox{\textwidth}{!}{
\begin{tabular}{lcccccc|c}
\toprule
\textbf{Model} & \textbf{RMSE} ↓ & \textbf{LogRMSE} ↓ & \textbf{MAE} ↓ & \textbf{MedAE} ↓ & \textbf{AbsRel} ↓ & \textbf{SqRel} ↓ & \textbf{ATE} ↓ \\
\midrule
Monodepth2~\cite{godard2019digging} & \underline{ 6.3138} & 0.1841 & \underline{ 4.3699} & \underline{ 3.1279} & 0.1501 & \underline{ 1.0168} &  2.7025\\
MonoViT~\cite{zhao2022monovit}  & 6.8544 & \textbf{0.1789} & 4.6146 & \textbf{3.0384} & \textbf{0.1473} & 1.0958  &  2.4298 \\
AF-SfM\textsubscript{\_pretrained}~\cite{shao2022self} & 17.3395 & 0.5216 & 12.0918 & 8.2881 & 0.4126 & 7.2491  &  - \\
AF-SfM\textsubscript{\_finetuned}~\cite{shao2022self} & 9.9636 & 0.2658 & 6.3069 & 3.4845 & 0.2165 & 3.7232  & 4.4841\\
IID-SfM\textsubscript{\_pretrained}~\cite{li2024image} & 17.6461 & 0.5269 & 12.3394 & 8.4590 & 0.4178 & 7.4915 & - \\
IID-SfM\textsubscript{\_finetuned}~\cite{li2024image} & 8.6977 & 0.2552 & 5.6891 & 3.5772 & 0.1996 & 1.9452 & 3.1358 \\
SHADeS~\cite{daher2025shades} & 7.1065 & 0.1845 & 4.7230 & 3.1341 & 0.1500 & 1.1437 & \textbf{1.2533}\\
Ours & \textbf{6.1246} & \underline{ 0.1830} & \textbf{4.3300} & 3.2582 & \underline{ 0.1475} & \textbf{1.0010} & \underline{1.9077}\\

\bottomrule
\end{tabular}}
\end{table}

%% file: TAB/Table_modality.tex
\begin{table}[t]
\centering
\caption{Quantitative comparison of depth estimation with models trained on Hyper-Kvasir and tested on the C3VD test set. Best results are in \textbf{bold} and second best results are in \underline{underlined}.}
\setlength{\tabcolsep}{0.5pt}
\renewcommand{\arraystretch}{1.2}
\resizebox{\textwidth}{!}{
\begin{tabular}{ccccccccccc|ccc}
\toprule
\textbf{Models} & 
\makecell{\textbf{Pose}\\\textbf{Edge}} & 
\makecell{\textbf{Pose}\\\textbf{Lum}} & 
\makecell{\textbf{Depth}\\\textbf{Edge}} & 
\makecell{\textbf{Depth}\\\textbf{Lum}} & 
\textbf{RMSE} ↓ & 
\makecell{\textbf{Log}\\\textbf{RMSE} ↓} & 
\textbf{MAE} ↓ & 
\textbf{MedAE} ↓ & 
\textbf{Abs Rel} ↓ & 
\textbf{Sq Rel} ↓ &
$\delta<1.25$ ↑ & 
$\delta<1.25^2$ ↑ &
$\delta<1.25^3$ ↑ \\
\midrule
Baseline & – & – & – & – & 6.2258 & 0.1845 & 4.3176 & 3.0993 & 0.1523 & 1.0360 & 0.7727 & 0.9741 & 0.9961 \\
\texttt{PE} & \checkmark & – & – & – & 6.2501& \underline{0.1824} & 4.3260 & \textbf{2.9790} & \underline{0.1505} &1.0610  & 0.7852 & 0.9665 & 0.9945\\
\texttt{PL} & – & \checkmark & – & – & 6.7658 & 0.2072 & 4.8068 & 3.5452 & 0.1719 & 1.2877 & 0.7318 & 0.9489 & 0.9910 \\
\texttt{DE} & – & – & \checkmark & – & 6.6625 & 0.2111 & 4.7625 & 3.4991 & 0.1755 & 1.3095 & 0.7338 & 0.9415 & 0.9891\\
\texttt{DL}&– & – & – & \checkmark & 6.7540 & 0.2005 & 4.7099 & 3.4705 & 0.1649 & 1.2652 & 0.7468 & 0.9576 & 0.9951 \\
\texttt{DEPL} & – & \checkmark & \checkmark & – & 6.9001 & 0.2105 & 4.9384 & 3.5381 & 0.1739 & 1.3404 & 0.7274 & 0.9483 & 0.9896 \\
\texttt{DLPE} & \checkmark & – & – & \checkmark &  \textbf{5.9500} & 0.1866 & \underline{4.2300} & 3.2132 & 0.1512 & \underline{1.0181} & 0.7689 & 0.9693 & 0.9974 \\
\texttt{DEPE} & \checkmark & – & \checkmark & – & 6.5529 & 0.1943 & 4.5099 & 3.0780 & 0.1558 & 1.1699 & 0.7623 & 0.9568 & 0.9901 \\
\texttt{DLPL} & – & \checkmark & – & \checkmark & \underline{6.0657} & \textbf{0.1751} & \textbf{4.2051} & \underline{2.9966} & \textbf{0.1413} & \textbf{0.9743} & 0.7992 & 0.9751 & 0.9980\\

\bottomrule
\end{tabular}}
\label{tab:depth_ablation_hk}
\end{table}

%% file: TAB/Table_mod_loss.tex
\begin{table}[t]
\centering
\caption{Depth and pose estimation results of models trained with Hyper-Kvasir and tested on C3VD test set with different combinations of structural priors and edge-guided loss. Best configuration per model group is in \textbf{bold}.}

\setlength{\tabcolsep}{2pt}
\renewcommand{\arraystretch}{1.2}
\resizebox{\textwidth}{!}{
\begin{tabular*}{\linewidth}{@{\extracolsep{\fill}}lcccc|cccccc}
\toprule
\multirow{2}{*}{\textbf{Model}} & 
\multirow{2}{*}{\textbf{Edge Loss}} & 
\multicolumn{3}{c|}{\textbf{Pose Estimation}} & 
\multicolumn{6}{c}{\textbf{Depth Estimation}} \\
\cmidrule(lr){3-5} \cmidrule(lr){6-11}
& & \textbf{ATE} ↓ & \textbf{RTE} ↓ & \textbf{ROT} & \textbf{RMSE} ↓ & \makecell{\textbf{Log}\\\textbf{RMSE} ↓} & \textbf{MAE} ↓ & \textbf{MedAE} ↓ & \textbf{AbsRel} ↓ & \textbf{SqRel} ↓ ↓ \\
\midrule

\multirow{2}{*}{{Monodepth2~\cite{godard2019digging}}} 
    & --  & 2.7025 & 0.2004 & 0.1250 & 6.3138 & 0.1841 & 4.3699 & 3.1279 & 0.1501 & 1.0168 \\ 
    & \checkmark  & \cellcolor{green!20}1.6293 & \cellcolor{red!20}0.2121 & \cellcolor{red!20}0.1286 & \cellcolor{red!20}6.8043 & \cellcolor{red!20}0.1991 & \cellcolor{red!20}4.9267 & \cellcolor{red!20}3.7038 & \cellcolor{red!20}0.1681 & \cellcolor{red!20}1.2604 \\
\midrule
\multirow{3}{*}{\texttt{DLPE}} 
    & --   & 2.5101 & 0.2005 & 0.1408 & 6.1246 & 0.1830 & 4.3300 & 3.2582 & 0.1475 & 1.0010 \\
    & \checkmark   & \cellcolor{red!20}2.5858 & \cellcolor{red!20}0.2015 & \cellcolor{red!20}0.1412 & \cellcolor{red!20}6.5946 & \cellcolor{green!20}0.1761 & \cellcolor{red!20}4.4031 & \cellcolor{green!20}2.9848 & \cellcolor{green!20}0.1388 & \cellcolor{red!20}1.0138 \\
    & \checkmark$^\dagger$ & \cellcolor{green!20}1.9077 & \cellcolor{red!20}0.2087 & \cellcolor{red!20}0.1496 & 6.1246 & 0.1830 & 4.3300 & 3.2582 & 0.1475 & 1.0010 \\
\midrule
\multirow{3}{*}{\texttt{DLPL}} 
    & --      & 3.1202 & 0.1953 & 0.1464 & 6.2562 & 0.1688 & 4.2677 & 2.9406 & 0.1341 & 0.9384 \\
    & \checkmark   & \cellcolor{green!20}2.1662 & \cellcolor{red!20}0.2050 & \cellcolor{green!20}0.1453 & \cellcolor{red!20}6.3535 & \cellcolor{red!20}0.1747 & \cellcolor{red!20}4.4400 & \cellcolor{red!20}3.2086 & \cellcolor{red!20}0.1426 & \cellcolor{red!20}1.0030 \\
    & \checkmark$^\dagger$ & \cellcolor{green!20}2.6802 & \cellcolor{red!20}0.2047 & \cellcolor{red!20}0.1458 & 6.2562 & 0.1688 & 4.2677 & 2.9406 & 0.1341 & 0.9384 \\
    



\bottomrule
\end{tabular*}}
\label{tab:mod_loss_ablation}
{\footnotesize\raggedright
\checkmark indicates that the edge-guided loss is jointly applied to both DepthNet and PoseNet during end-to-end training. \checkmark$^\dagger$ denotes stage-wise training, where the edge-guided loss is only introduced when fine-tuning PoseNet, with DepthNet frozen. \textcolor{red}{Red-shaded} cells indicate that the metric value is worse compared to the baseline (“--”) in the same block, while \textcolor{green!50!black}{green-shaded} cells indicate improvement.\par

}
\end{table}

%% file: TAB/Table_temporal.tex
\begin{table*}[t]
\centering
\caption{Quantitative depth estimation results across models, datasets, and frame intervals. All models are trained on either C3VD or Hyper-Kvasir and always tested on C3VD test data.}
\label{tab:temporal}
\setlength{\tabcolsep}{3pt}
\renewcommand{\arraystretch}{1.2}
\resizebox{\textwidth}{!}{
\begin{tabular}{lll|cccccc|ccc}
\toprule
\textbf{Model} & \textbf{Training Dataset} & \textbf{Interval} & \textbf{RMSE} $\downarrow$ & \textbf{RMSE log} $\downarrow$ & \textbf{MAE} $\downarrow$ & \textbf{MedAE} $\downarrow$ & \textbf{Abs Rel} $\downarrow$ & \textbf{Sq Rel} $\downarrow$ & \textbf{$\delta<1.25$} $\uparrow$ & \textbf{$\delta<1.25^2$} $\uparrow$ & \textbf{$\delta<1.25^3$} $\uparrow$ \\
\midrule
\parbox[t]{2mm}{\multirow{8}{*}{\rotatebox[origin=c]{90}{\textbf{Monodepth2~\cite{godard2019digging}}}}} 
    & \multirow{4}{*}{C3VD}
        & 1 & 15.1642 & 0.4196 & 10.1482 & 6.3000 & 0.3047 & 4.7856 & 0.4516 & 0.7353 & 0.8783 \\
    &   & 10 & \textbf{7.5960} & \textbf{0.2215} & \textbf{5.0917} & \textbf{3.4709} & \textbf{0.1732} & \textbf{1.4608} & \textbf{0.7161} & \textbf{0.9253} & \textbf{0.9827} \\
    &   & 20 & 7.7644 & 0.2473 & 5.4523 & 3.9334 & 0.2002 & 1.7340 & 0.6668 & 0.8976 & 0.9725 \\
    &   & 30 & 7.7282 & 0.2403 & 5.2392 & 3.6111 & 0.1878 & 1.6480 & 0.6919 & 0.8956 & 0.9696 \\
\cmidrule(lr){2-12}
    & \multirow{4}{*}{Hyper-Kvasir}
        & 1 & \textbf{6.3138} & \textbf{0.1841} & \textbf{4.3699} & \textbf{3.1279} & \textbf{0.1501} & \textbf{1.0168} & \textbf{0.7754} & \textbf{0.9720} & \textbf{0.9956} \\
    &   & 10 & 40.5039 & 1.3728 & 32.2524 & 26.0333 & 1.5865 & 93.1729 & 0.1331 & 0.2595 & 0.3706 \\
    &   & 20 & 17.4583 & 0.5306 & 12.2546 & 8.6783 & 0.4248 & 7.4185 & 0.3636 & 0.6225 & 0.7901 \\
    &   & 30 & 17.4583 & 0.5306 & 12.2546 & 8.6783 & 0.4248 & 7.4185 & 0.3636 & 0.6225 & 0.7901 \\
\midrule
\parbox[t]{2mm}{\multirow{8}{*}{\rotatebox[origin=c]{90}{\textbf{MonoViT~\cite{zhao2022monovit}}}}} 
    & \multirow{4}{*}{C3VD} 
        & 1 & 15.9045 & 0.4543 & 10.7178 & 6.7776 & 0.3394 & 5.5865 & 0.4373 & 0.7044 & 0.8527 \\
    &   & 10 & \textbf{6.2371} & \textbf{0.1625} & \textbf{3.8398} & \textbf{2.3075} & \textbf{0.1203} & \textbf{0.8766} & \textbf{0.8427} & \textbf{0.9753} & \textbf{0.9929} \\
    &   & 20 & 6.5954 & 0.1797 & 4.1504 & 2.4064 & 0.1350 & 1.0694 & 0.8139 & 0.9442 & 0.9902 \\
    &   & 30 & 6.4839 & 0.1804 & 4.0457 & 2.4405 & 0.1397 & 1.1434 & 0.8137 & 0.9415 & 0.9900 \\
\cmidrule(lr){2-12}
    & \multirow{4}{*}{Hyper-Kvasir}
        & 1 & \textbf{6.8544} & 0.1789 & \textbf{4.6146} & \textbf{3.0384} & \textbf{0.1473} & \textbf{1.0958} & \textbf{0.7884} & \textbf{0.9788} & \textbf{0.9982} \\
    &   & 10 & 30.9196 & \textbf{1.1751} & 24.6074 & 21.0623 & 0.9914 & 35.7517 & 0.1471 & 0.2997 & 0.4428 \\
    &   & 20 & 30.8911 & 1.0684 & 23.9879 & 19.5039 & 1.1202 & 51.3599 & 0.1968 & 0.3551 & 0.4864 \\
    &   & 30 & 34.5043 & 1.1523 & 26.7642 & 21.4713 & 1.2017 & 58.8600 & 0.1739 & 0.3336 & 0.4751 \\
\midrule
\parbox[t]{2mm}{\multirow{8}{*}{\rotatebox[origin=c]{90}{\textbf{AF-SfMLearner~\cite{shao2022self}}}}}
    & \multirow{4}{*}{C3VD}
        & 1 & \textbf{16.8063} & \textbf{0.4929} & \textbf{11.3895} & \textbf{7.4673} & \textbf{0.3737} & \textbf{6.6151} & \textbf{0.4128} & \textbf{0.6786} & \textbf{0.8312} \\
     &  & 10 & 17.1426 & 0.5008 & 11.7025 & 7.5752 & 0.3847 & 6.8968 & 0.3962 & 0.6698 & 0.8239 \\
     &  & 20 & 17.2341 & 0.5128 & 11.8870 & 7.8705 & 0.3918 & 6.9788 & 0.3735 & 0.6483 & 0.8199 \\
     &  & 30 & 17.4512 & 0.5248 & 12.3266 & 8.5111 & 0.4245 & 7.6151 & 0.3596 & 0.6349 & 0.8001 \\
\cmidrule(lr){2-12}
    & \multirow{4}{*}{Hyper-Kvasir}
        & 1 & 9.9727 & \textbf{0.2660} & 6.3114 & \textbf{3.4862} & \textbf{0.2166} & 3.7265 & \textbf{0.6719} & \textbf{0.8900} & 0.9591 \\
     &  & 10 & 11.5691 & 0.3057 & 7.1733 & 4.1365 & 0.2540 & 4.5569 & 0.6146 & 0.8672 & 0.9556 \\
     &  & 20 & 10.1226 & 0.2833 & 6.4429 & 3.8149 & 0.2223 & \textbf{2.7322} & 0.6576 & 0.8663 & 0.9534 \\
     &  & 30 & \textbf{9.9544} & 0.2740 & \textbf{6.2987} & 3.8747 & 0.2240 & 3.2430 & 0.6713 & 0.8829 & \textbf{0.9625} \\
\midrule
\parbox[t]{2mm}{\multirow{8}{*}{\rotatebox[origin=c]{90}{\textbf{IID-SfMLearner~\cite{li2024image}}}}}
    & \multirow{4}{*}{C3VD}
        & 1 & 14.3235 & 0.3844 & 9.3650 & \textbf{5.4199} & \textbf{0.2766} & 4.1887 & \textbf{0.4989} & 0.7773 & 0.9040 \\
     &  & 10 & 13.8868 & \textbf{0.3823} & \textbf{9.1635} & 5.5565 & 0.2827 & \textbf{4.1115} & 0.4973 & \textbf{0.7853} & \textbf{0.9094} \\
     &  & 20 & 14.3038 & 0.3988 & 9.5545 & 6.0755 & 0.2996 & 4.4211 & 0.4753 & 0.7667 & 0.9006 \\
     &  & 30 & \textbf{13.7964} & 0.3858 & 9.1892 & 5.8469 & 0.2967 & 4.1495 & 0.4856 & 0.7791 & 0.9056 \\
\cmidrule(lr){2-12}
    & \multirow{4}{*}{Hyper-Kvasir}
        & 1 & 8.6977 & 0.2552 & 5.6891 & 3.5772 & 0.1996 & 1.9452 & 0.6656 & 0.8773 & 0.9624 \\
     &  & 10 & \textbf{8.4079} & \textbf{0.2319} & \textbf{5.3326} & \textbf{3.2531} & 0.1829 & \textbf{1.6545} & \textbf{0.7237} & 0.9101 & 0.9767 \\
     &  & 20 & 8.8540 & 0.2357 & 5.6743 & 3.5312 & \textbf{0.1821} & 1.7427 & 0.7007 & \textbf{0.9103} & \textbf{0.9807} \\
     &  & 30 & 9.1189 & 0.2465 & 5.8221 & 3.4976 & 0.1900 & 1.9096 & 0.6988 & 0.8968 & 0.9701 \\
\bottomrule
\end{tabular}}
\end{table*}